\documentclass{article}
\usepackage{amssymb}

\usepackage[final]{corl_2025} % Uncomment for the camera-ready ``final'' version.
\usepackage{wrapfig}
\usepackage{xspace}
\usepackage{graphicx} % Comment this line out if you need a4paper
\usepackage{xcolor}
\usepackage{float} 
\usepackage{hyperref}
\usepackage{enumitem}
\usepackage{booktabs}
\usepackage{float}
\usepackage{amsmath} 
\usepackage{cleveref}
\usepackage{soul}
\definecolor{myGreen}{HTML}{7aa796}
\definecolor{StanfordRed}{rgb}{0.549, 0.082, 0.082}
\usepackage[dvipsnames]{xcolor}
\hypersetup{
    colorlinks=true,
    linkcolor=cyan,
    filecolor=magenta,      
    urlcolor=StanfordRed
    }
\newcommand{\ours}{\textbf{\textsc{1001 Demos}}}

\title{ One Demo is Worth a Thousand Trajectories: Action-View Augmentation for Visuomotor Policies
}

\author{
  {\bfseries Chuer Pan$^{1}$ \quad Litian Liang$^{1}$ \quad Dominik Bauer$^{2}$} \\[1ex]
  {\bfseries Eric Cousineau$^{3}$ \quad Benjamin Burchfiel$^{3}$ \quad Siyuan Feng$^{3}$ \quad Shuran Song$^{1}$} \\[1.5ex]
  {\normalfont\normalsize $^{1}$Stanford University, $^{2}$Columbia University, $^{3}$Toyota Research Institute} \\
  \url{https://chuerpan.com/1001-demos.github.io/}
}
\begin{document}
\maketitle
\begin{abstract}
Visuomotor policies for manipulation have demonstrated remarkable potential in modeling complex robotic behaviors, yet minor alterations in the robot’s initial configuration and unseen obstacles easily lead to out-of-distribution observations. Without extensive data collection effort, these result in catastrophic execution failures.
In this work, we introduce an effective data augmentation framework that generates visually realistic fisheye image sequences and corresponding physically feasible action trajectories from real-world eye-in-hand demonstrations, captured with a portable parallel gripper with a single fisheye camera. 
We introduce a novel Gaussian Splatting formulation, adapted to wide FoV fisheye cameras, to reconstruct and edit the 3D scene with unseen objects. We utilize trajectory optimization to generate smooth, collision-free, view-rendering-friendly action trajectories and render visual observations from corresponding novel views. Comprehensive experiments in simulation and the real world show that our augmentation framework improves the success rate for various manipulation tasks in both the same scene and the augmented scene with obstacles requiring collision avoidance. %\href{https://sites.google.com/view/1001-demos-corl?usp=sharing}{website} 
\end{abstract}

%they easily go out of distribution when encountering unseen visual observations and thereby exhibit brittleness in real-world applications. M
\section{Introduction}

\begin{wrapfigure}{r}{0.65\textwidth}
\vspace{-15mm}
  \begin{center}
    \includegraphics[width=0.99\linewidth]{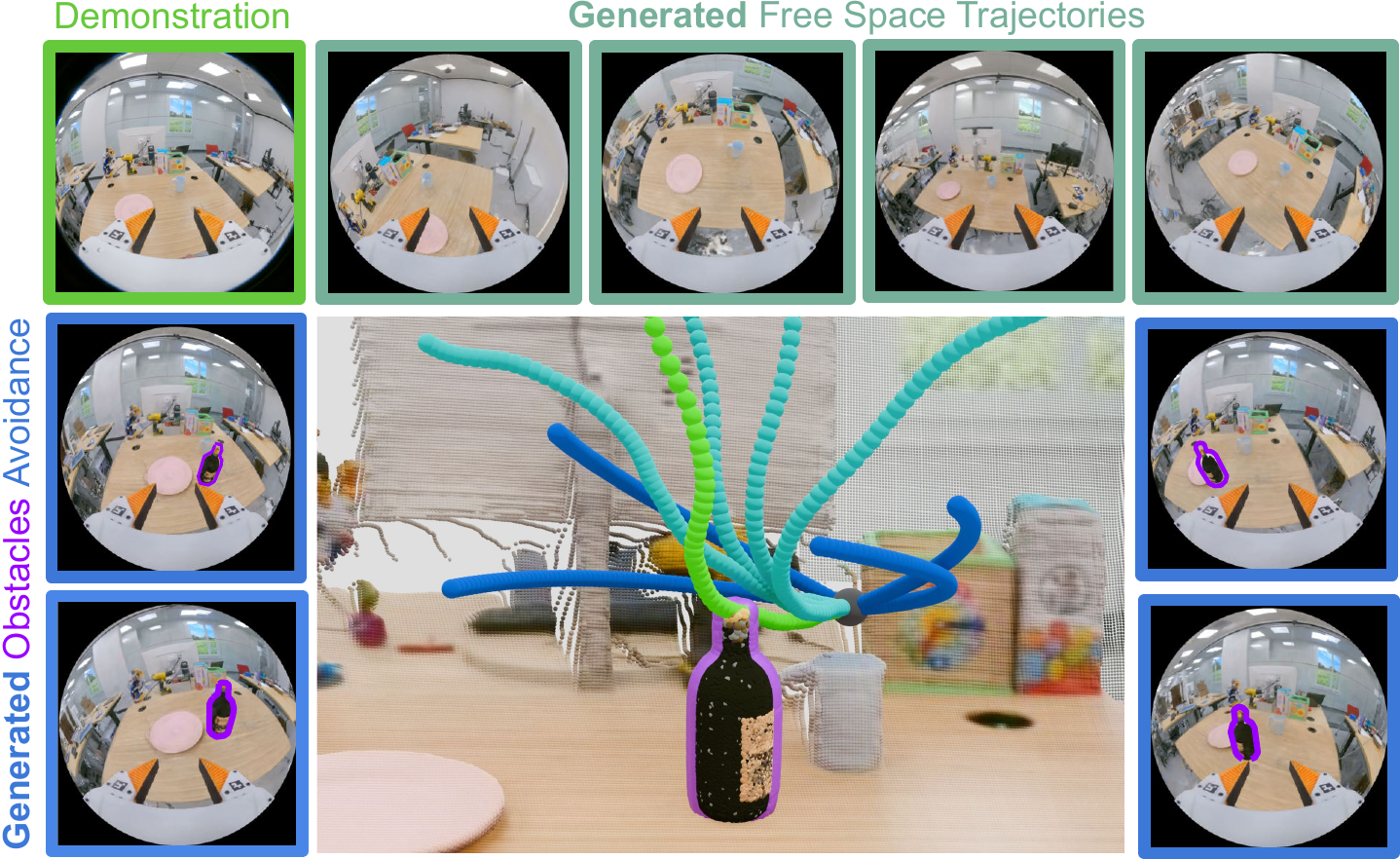} %1001 Teaser.pdf}
  \end{center}
  % https://docs.google.com/drawings/d/1ATdHLmdSJjqKztAh39FyipNNWk6G17pn2iJZX7bOKa8/edit?usp=sharing
\vspace{-2mm}
    \caption{\textbf{\ours.} From a single \textcolor[HTML]{57c027}{human demonstration} (e.g., picking up the blue mug), our approach generates valid training trajectories with \textcolor[HTML]{3eb0b0}{large spatial variance} and \textcolor[HTML]{025db3}{augmented obstacles}, while respecting action-view consistency, 3D collision and contact dynamics constraints.      } 
     \label{fig:teaser}
     \vspace{-3mm}
\end{wrapfigure}

% why Data augmentation 
%human demonstrators often have to give policies spatial robustness by performing spatial augmentation of the same skill
Visuomotor policies trained through imitation learning~\cite{chi2023diffusion,zhao2023learning,shafiullah2022behavior} enable complex robot behaviors but often remain brittle: minor changes in the robot's initial configuration or the objects in the scene may yield out‐of‐distribution (OOD) observations, cascading into OOD states, and resulting in compounded execution errors that cause task failures, hindering robot performance~\cite{ross2011reduction,team2024octo,mirchandani2024so}.
To improve policies' \textbf{spatial robustness}, human demonstrators have to repeatedly demonstrate the same skill on identical objects numerous times under different spatial configurations~\cite{chi2024universal}. While effective, this manual process is tedious and costly. We address this by introducing an effective data augmentation framework that improves the spatial robustness of visuomotor policies by automatically generating additional real-world robot trajectory data from existing human demonstrations, thereby expanding data spatial coverage, without exhaustive manual collection.

While data augmentation is a standard procedure in other domains, such as computer vision~\cite{shorten2019survey}, augmenting real-world robot manipulation data presents a set of unique challenges: 
\begin{itemize}[leftmargin=3mm]
    \item \textbf{Maintaining Action-View Consistency.} Robot policy learning requires paired observation and action as data. 
    Hence, the data argumentation algorithm need to increase both visual and action diversity and critically maintain the consistency between these two. 
    %Visual behavior cloning methods leverage sequences of corresponding observation and action pairs as training data. To produce robot-executable manipulation augmentation data thus requires us to produce both \textit{visually realistic} visual observations and \textit{physically feasible} actions associated with the generated visual observations.
    
    % Need to improve both visual diversity and action. Need to make sure the augmentation between action and visual data is consistent. 
    
    \item \textbf{Respecting Physical Constraints of Actions.} The augmented action data needs to obey physical constraints, including both 3D collision constraints and object contact dynamics. 
    
    \item \textbf{Maximizing Visual Coverage from Limited Demonstrations.} Real-world robot demonstration data is limited in terms of view coverage. Therefore, to make effective data argumentation possible, we need to make every collected trajectory demo count by maximizing their visual coverage from the same number of views during data collection.
\end{itemize}

% what we do. 
% To address these challenges, we propose \ours, a data augmentation technique with the following key design elements:
To address these challenges, we propose \ours; a data augmentation technique featuring the following key designs: 
% We propose a \emph{trajectory-level action–view augmentation} algorithm that first reconstructs the environemnt, then leverages trajectory optimization to generate physically feasible action trajectories, and uses 3D Gaussian Splatting for novel-view synthesis to generate corresponding visually and spatially realistic observations, to get consistent observation–action pairs.

% We propose a \emph{trajectory-level action–view augmentation} algorithm that first reconstruct the scene, generate physically feasible, collsion-free action trajectories via trajectory optimization , and renders its spatially matching visual observations via 3D Gaussian Splatting novel view sythesis, yielding spatially consistent, visually realistic, and physically feasible observation–action demonstration episodes.

\begin{itemize}[leftmargin=3mm]
    \item To generate spatially consistent observation-action pairs, we propose a \emph{trajectory-level action–view augmentation} algorithm that first reconstructs the scene, then generates physically feasible, collision-free action trajectories via trajectory optimization, to finally render the spatially matching observations via (editable) 3D Gaussian Splatting (3DGS); yielding spatially consistent, visually realistic, and physically feasible observation–action demonstration trajectory episodes. 
    %Different from single-step argumentation~\cite{zhou2023nerf, zhang2024diffusion}, our action generation step considers trajectory level smoothness, and allows larger spatial variance, and spatilaly consistent scene editing.
    
    \item To obey the 3D collision constraints in 3DGS scenes with edited obstacles, we propose a \textit{collision-aware} action generation module that uses trajectory optimization to create smooth, collision-free, and diverse action trajectories beyond the demonstrated action distribution, allowing the resulting visuomotor policies to learn collision-avoidance behaviors. 
    To obey object contact dynamics, we propose a \textit{contact-aware} augmentation for automatic contact event detection, which only perform augmentation before contact events, preserving the contact dynamics in the original demo.

    %This enables policies to train from data that handle collision avoidance by augmenting 3D obstacles (from Objaverse~\cite{deitke2023objaverse}) in the scene and augment corresponding collision-free trajectory data via trajectory optimization. 

     \item To maximize visual coverage during data collection, our system uses an ultra-wide fish-eye camera. However, while this drastically increases the field-of-view of the observations, this non-standard camera configuration requires us to extend the 3D Gaussian Splatting formulation by introducing a \textit{fisheye ray sampler} in the rendering step.
   
\end{itemize}

% summarize the results:
Our experiments in simulation and the real world validate the effectiveness of our action-view data augmentation approach. We show that the proposed free-space data augmentation improves the manipulation policies' performance in simulation on the RoboMimic~\cite{mandlekar2021matters} benchmark. Moreover, the proposed collision-aware data augmentation improves real-world manipulation policies robustness against unseen obstacles in pick-and-place and non-prehensile tasks, leading to an improved success rate when compared to policies that are trained without \ours.

\section{Related Work}
% In the following, we discuss data augmentation techniques in policy learning; from purely visual augmentation, over exploiting state information, to closely related methods that jointly augment visual observations and desired actions.

\textbf{Image Augmentation for Visual Invariance.}
Robustness to visual variation -- appearance, illumination, viewpoint -- has been extensively studied~\cite{shorten2019survey}. Common augmentations include color jittering~\cite{chi2023diffusion}, image filtering~\cite{hansen2021generalization}, and cropping~\cite{laskin2020reinforcement,yarats2021image}. Image editing using generative models further enables object-level modifications~\cite{chen2024semantically,yu2023scaling,black2023zero}, embodiment swapping~\cite{chen2024rovi}, and viewpoint interpolation with embodiment transfer~\cite{chen2025tool}.
Yet, such approaches are not able to augment the desired robot trajectory accordingly, and are thus limited to global appearance, background object, or minor viewpoint changes. 
By contrast, our method synthesizes visually realistic, multi‐view observations via 3DGS and produces physically feasible action trajectories through trajectory optimization.

\textbf{State-based Data Augmentation.}
State-based augmentation disentangles raw visual inputs from policy observations by varying scene configurations and adjusting trajectories. For example, Florence et al.~\cite{florence2019self} inject noise into keypoint-based state representations to mitigate cascading errors. Learned‐dynamics methods with continuity constraints guide policies back to expert states~\cite{ke2023ccil,deshpande2024data}. Simulation environments further ensure the validity of such augmentations~\cite{mitrano2022data}.
However, these methods defer visual invariance to state estimation and rely on a dynamics model -- simulated or learned -- to ensure physical consistency. Instead, our approach jointly augments visual inputs and action trajectories in a realistic manner, producing diverse, obstacle‐avoiding demonstrations that yield policies robust to out‐of‐distribution viewpoints and capable of obstacle avoidance.

\textbf{Visual-Action Augmentation.}
Augmenting visual observations \textit{and} actions enables joint variation of scene configurations and robot behaviors. Prior work generates third‐person pinhole augmentations, either in simulation~\cite{mandlekar2023mimicgen,jiang2024dexmimicgen}, via novel‐view synthesis~\cite{yang2025novel,tian2024view} and novel 3D configuration synthesis~\cite{pan2023tax} in the real world, or adapt egocentric pinhole observations via NeRF/3DGS and MPC~\cite{tagliabue2024tube,low2025sous}. Zhou et al.~\cite{zhou2023nerf} (using NeRF) and Zhang et al.~\cite{zhang2024diffusion} (using diffusion) focus on single‐step pinhole image-action pairs but can neither handle wide FoV nor produce trajectory‐level, obstacle‐avoiding data. 
MimicGen~\cite{mandlekar2023mimicgen} and follow-ups~\cite{hoque2024intervengen, jiang2024dexmimicgen, garrett2024skillmimicgen} augment third-person demonstrations but rely on costly on-robot rollouts to obtain in-domain visuals for real-world deployment. Concurrent work~\cite{robosplat, xue2025demogen} enable trajectory-level visual-action augmentation, but both are restricted to demonstrations from \textit{static, third-person, pinhole} cameras. Yang et al. use 3DGS to edit robot, object, and background appearances but does not allow obstacle avoidance augmentation; Xue et al. focus on visuomotor policies with point cloud inputs, using point cloud editing to produce obstacle-aware augmentations. In contrast, we integrate a fisheye ray sampler into 3DGS to elegantly handle \textit{fisheye} images -- preserving 3DGS' speed and editability while enabling \textit{trajectory‐level}, \textit{obstacle‐avoiding} for \textit{eye-in-hand} observation-action demonstrations; creating robust visuomotor policies across diverse camera viewpoints and obstacle avoidance behaviours.

\section{The 1001 Demos Framework}
We introduce \ours, an offline data augmentation framework for visuomotor policies. From a single real-world%egocentric manipulation 
task demonstration, using a portable manipulation-data collection device equipped with a fisheye camera,  our augmentation technique generates~\ours~of \textit{visually realistic} image sequences for \textit{physically feasible} action trajectories. % DB: this is actually the best formulation so far of this connection; consistency aspect is still a bit implicit but workshopping this sentence might get you there

\begin{figure}[h]
    \centering
    \vspace{-1.5ex}
    \includegraphics[width=\linewidth]{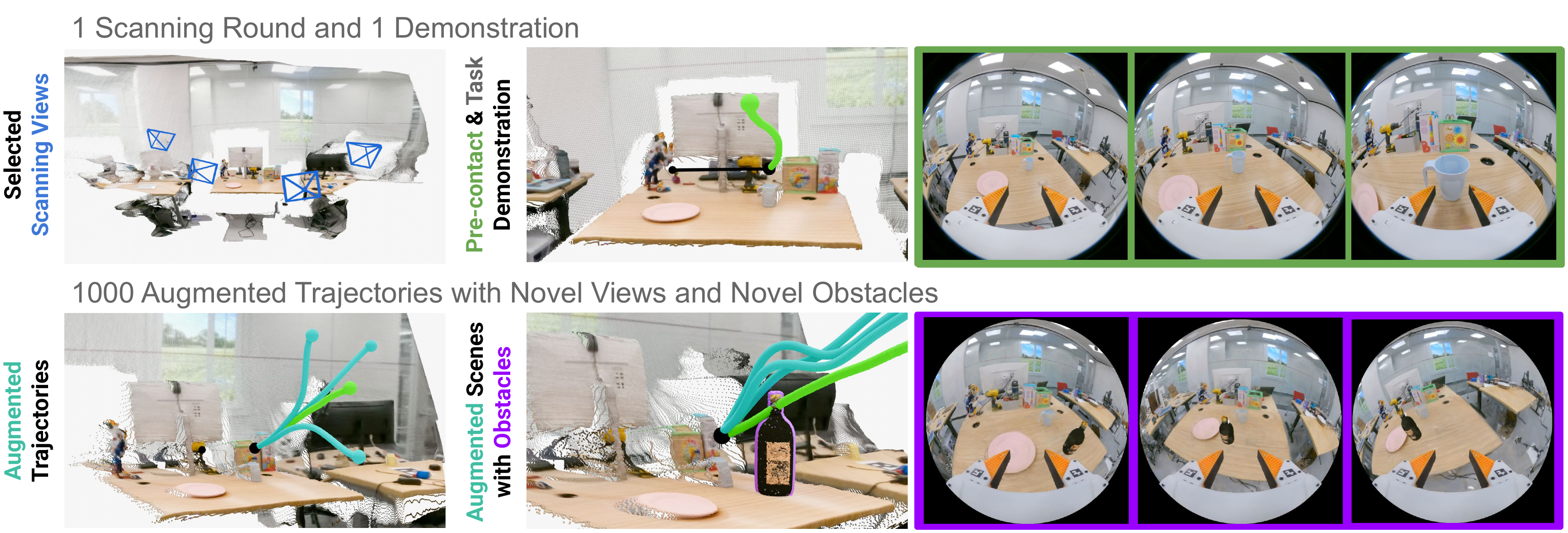} % https://docs.google.com/drawings/d/177WnpQop5BXm7Qb89bFZ5TOAoU14-wHX6Z6eR2EskXk/edit?usp=sharing
    \vspace{-2.5ex}
    \caption{\textbf{\ours~Overview.} From an initial mapping run, we reconstruct the 3D scene point cloud for easy trajectory planning and a fisheye 3DGS scene for fast novel view rendering (\S \ref{sec:method-scene-reconstruction}). Given a single demonstration video (green), we optimize additional physically feasible action trajectories (\S \ref{sec:method-demo-action}) and render the corresponding visually consistent fisheye-image observations (\S \ref{sec:method-demo-view}), thereby generating thousands of diverse action-view demonstrations from a single real-world demonstration.}
     
    \label{fig:method-flow}
\end{figure}

As illustrated in~Fig.~\ref{fig:method-flow}, given a fisheye video pair -- from scene scanning and task demonstration, \ours~is able to generate (1) demonstrations with vastly different initial configurations and (2) collision-avoiding demonstrations with obstacles added through scene editing. \ours~achieves this via three modules.
First, we use fisheye image sequences to reconstruct %high-fidelity 
a 3D scene point cloud for motion planning and a \textit{fisheye} 3D Gaussians representation of the scene for novel view rendering (\S \ref{sec:method-scene-reconstruction}). % DB: I don't quite get the capitalization of Fisheye/fisheye depending on the context here - are we trying to coin the term Fisheye 3DGS? I thought that was already used by another method - so that might actually be confusing - resolved
Second, given the extracted scene point cloud and the original demonstration trajectory, we generate smooth and collision-free trajectories in the same scene, starting from different initial camera poses (\S \ref{sec:method-demo-action}) and render the corresponding novel observations (\S \ref{sec:method-demo-view}).
Finally, with the generated free-space and obstacle-avoiding demonstrations, we train visuomotor policies on the original expert-collected demonstration and the generated demonstrations. %with Diffusion Policy~\cite{chi2023diffusion}
(\S \ref{sec:method-policy}), enabling downstream robot policies that gracefully handle unseen initial configurations and avoid novel obstacles in the scene.

% \ours ~first parses the task demonstration into pre-contact and post-contact segments with our contact detection module (Sec \ref{sec:method-contact-detection}). 

% Next, with the mapping and pre-contact fisheye image sequences, we produce a 3D point cloud of the scene with COLMAP and, subsequently a 3D Gaussian representation of the scene with our custom Fisheye 3DGS formulation (Sec \ref{sec:fisheye-3dgs}). 

% For each pre-contact demonstration, we sample new initial poses by significantly perturbing the original start pose (up to $90^{\circ}$) and utilize trajectory optimization (Sec \ref{sec:method-traj-opt-free-space}) to generate smooth 6D trajectories that funnel back and reconnect to the post-contact demo trajectory, accompanied with rendering-friendly camera views. 

\subsection{3D Scene Reconstruction from Eye-in-Hand Fisheye Video}
\label{sec:method-scene-reconstruction}

We use UMI~\cite{chi2024universal} and its demonstration dataset format for hand-held data collection, and Diffusion Policy~\cite{chi2023diffusion} for policy learning. Specifically, we assume that each dataset $\mathcal{D} = \{d\}_{N}$ consists of $N$ data episodes. Each episode $d = (o, a)$ is composed of a sequence of visual observations $o \triangleq \{o_{fish}\}$ as eye-in-hand fisheye RGB images, $o_{fish}$, and a sequence of action, $a \triangleq \{ a_{ee}, a_{gp}\}$. Where each action $a$ is composed of a 6D end-effector pose, $a_{ee} \in \mathbf{SE}(3)$, and a gripper-opening width, $a_{gp} \in \mathbb{R}$. Given a video pair, collected during scene scanning and task demonstration, we use these fisheye image sequences for 3D reconstruction to produce a %high-fidelity % DB: what makes this a high-fidelity point cloud? do you mean dense?
3D scene point cloud for motion planning and a Fisheye 3D Gaussian representation of the scene for novel view rendering. % DB: is this sentence repeated directly from the section intro? if so, please rephrase this a bit at least

\textbf{Scene Point Cloud Reconstruction \& Contact Detection.}
\label{sec:method-scene-reconstruction-pcd}
We leverage COLMAP~\cite{schoenberger2016sfm} to reconstruct high-fidelity 3D point clouds from the fisheye image sequences captured during scene scanning. We then split each demonstration into pre-contact and post-contact phases by finding the first frame where the gripper exceeds a collision threshold with the reconstructed point cloud. 

\begin{wrapfigure}{r}{0.4\textwidth}  
%\vspace{-5mm}
    \centering
\includegraphics[width=0.99\linewidth]{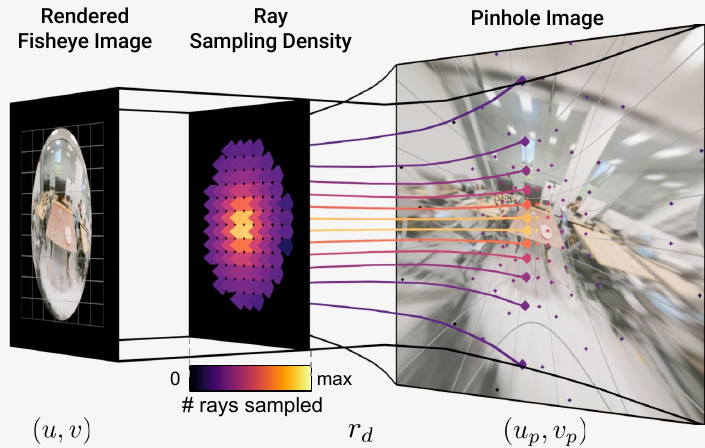} % https://docs.google.com/drawings/d/1nICJwYR9SxZ_dJvhtbUg6ZadfzrHmOz3oGq8kiNOLY4/edit?usp=sharing 
    \vspace{-6mm}
    \caption{ \textbf{Fisheye 3DGS.}
    We propose Fisheye-3DGS, using a ray sampler that accounts for fisheye distortion. Sampling density adapts to pixel location, allocating more rays to the image center than the periphery for better rasterization quality.
    }
    \label{fig:fisheye-gs}
    \vspace{-5mm}
\end{wrapfigure}

\textbf{Fisheye 3D Gaussians.}
\label{sec:method-scene-reconstruction-fisheye-gs}
A critical design choice enabling fast rasterization, 3DGS~\cite{kerbl20233d} tiles pinhole image into $16\times16$ pixel patches and uses 256-thread cuda blocks per tile, one thread per pinhole ray. To elegantly handle fisheye images, we replace the original ray sampler with a KB8-based~\cite{kannala2006generic} fisheye ray sampler. As shown in Fig.\ref{fig:fisheye-gs}, for each fisheye pixel \((u,v)\), compute its ray direction $r_d \;=\;\mathrm{KB8}(u,v)$~\cite{kannala2006generic}.
then project \(r_d\) through the camera intrinsics \(K\) to obtain the pinhole coordinates $(u_p, v_p)=K\,r_d$, thus associating each fisheye ray with its 3D-Gaussian splat location on the 2D image plane. We redistribute tile assignments from pinhole to fisheye rays, partition fisheye rays into the original 256-ray-per-tile layout to preserve the CUDA block–thread structure for fast rasterization while accurately modeling fisheye distortion. We optimize the fisheye 3DGS with pixel-wise losses from~\cite{kerbl20233d} between rendered and ground-truth fisheye images.

\subsection{Action Generation via Trajectory Optimization}\label{sec:method-demo-action}
\label{sec:method-traj-opt}

Utilizing the extracted 3D scene point cloud from Sec~\ref{sec:method-scene-reconstruction-pcd} % DB: Sec with or without "."? check conference style guide and define \cref or \Cref accordingly - use that throughout for sections, figures, tables
and the input demonstration trajectory, we employ trajectory optimization to generate two types of novel trajectories: (1) \textit{free space} demonstration trajectories in the same scene with different starting poses, sampled randomly % DB: I feel this under-sells you sampling strategy; it's not really random but on some manifold, a spherical circle constrained by the quaternion cone (sounds fancier, doesn't it? ;) )
in the free space of the scene; (2) \textit{obstacle-avoiding} demonstration trajectories that are planned around the added obstacle point cloud in the scene. 

\textbf{Trajectory Optimization Formulation.} 
Given a sequence of 6D camera poses, $O_\mathrm{ee}\triangleq\{o_\mathrm{ee}^{m}\}_{m=1}^{H} \subset \textbf{SE}(3)$ -- % DB: define - what is H (some horizon)? what is T (below)? why is m \neq k? why is k used in the eq below but m not? do we need this time indexing here?
extracted from the pair of scene scanning and task demonstration --, the 3D point cloud of the task scene, $P_\mathrm{scene}\in \mathbb{R}^{N_\mathrm{scene}\times3}$, and the start and end poses specified as $x_\mathrm{init} \in \textbf{SE}(3)$ and $x_\mathrm{goal} \in \textbf{SE}(3)$ (chosen as the pre-contact pose here), respectively. Provided with a trajectory initialization, $X \triangleq \{x^k\}_{k=1}^{T} \subset \textbf{SE}(3)$, we consider the following trajectory optimization problem,

% DB: afaik equations should always be numbered -- i.e., use the \begin{equation} instead -- so you and others can refer to it by a number. also, give this thing some variable like \mathcal{L} 
\[
\begin{aligned}
\operatorname*{argmin}_{\{x_k\}_{k=1}^{T}} \quad & \mathcal{L}_\mathrm{funnel}(X,  O_\mathrm{ee})) + \mathcal{L}_\mathrm{collision}(X, \text{tsdf}(P_\mathrm{scene})) + \mathcal{L}_\mathrm{render}(X, O_\mathrm{ee}) + \mathcal{L}_\mathrm{smooth}(X), \\
\text{subject to} \quad & x^1=x_\mathrm{init}, x^{T}=x_\mathrm{goal},  X \cap \mathrm{convhull}(P_\mathrm{obstacle})= \emptyset ,
\end{aligned}
\]
% DB: if you write an equation like this, you should treat it like text - by which I mean, you need to set "," and "." (as needed).
where $P_\mathrm{obs} \in \mathbb{R}^{N_\mathrm{obs}\times3}$ is the point cloud of the augmented obstacle added for obstacle-avoiding trajectory generation (Sec \ref{sec:method-traj-opt-obstacle}). $\mathrm{convhull}(\cdot)$ retrieves the convex hull from a point cloud, $\mathrm{tsdf}(\cdot):\mathbb{R}^3 \rightarrow \mathbb{R}$ is a truncated signed distance function (TSDF) which maps a 3D coordinate to a scalar distance. $x_\mathrm{init}$ is sampled within a $\theta$ quaternion cone around each original viewpoint, with $r$ as the radius of the quaternion sphere.

\textbf{Free-Space Trajectory Generation.}
\label{sec:method-traj-opt-free-space}

\begin{wrapfigure}{r}{0.4\textwidth}  
    \centering
    \vspace{-5mm}
    \includegraphics[width=\linewidth]{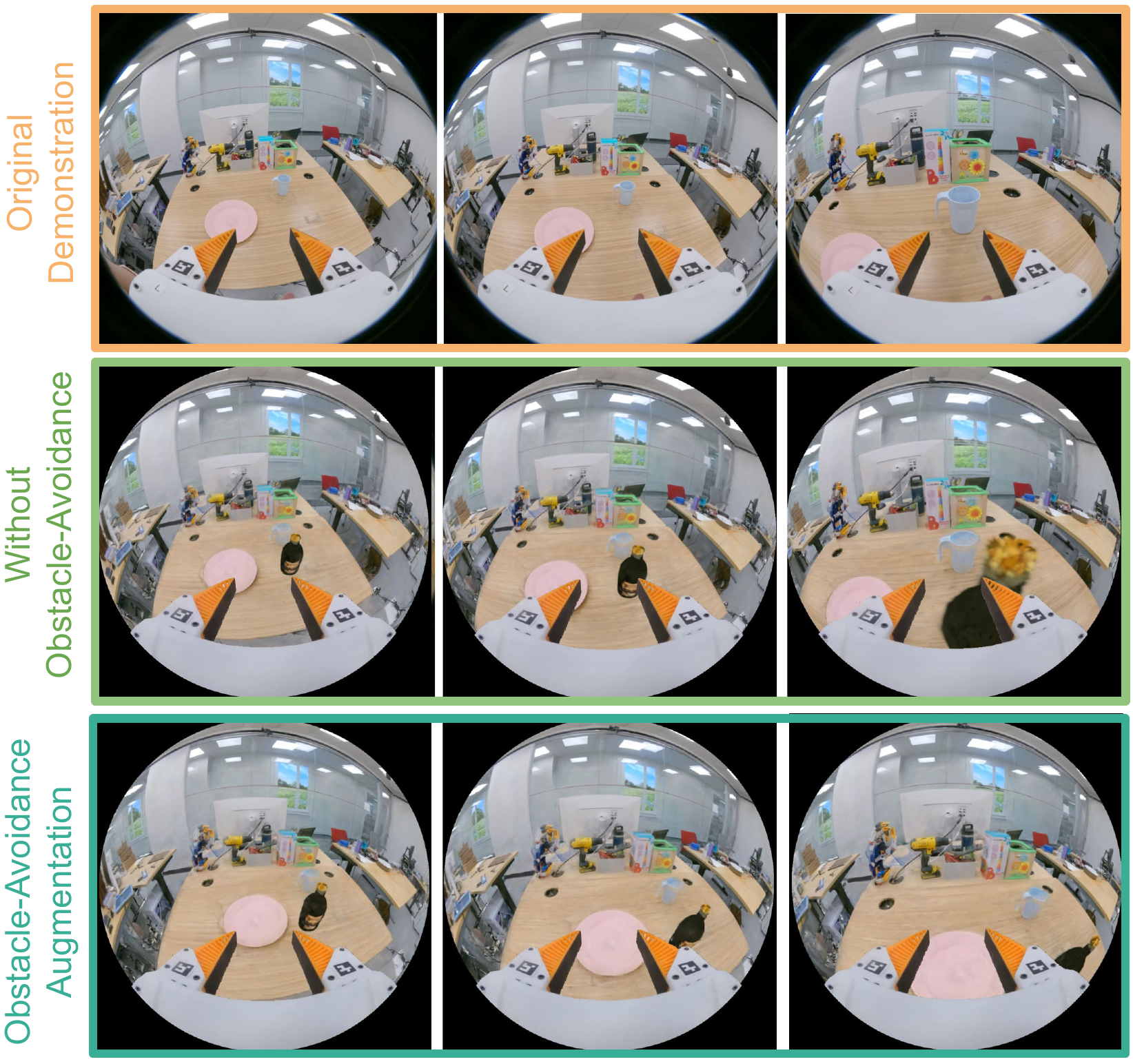} % https://docs.google.com/drawings/d/1f0akTNsbzF0gdQgmobVUiCCFn2fVpb6GiuLbvFE5gZI/edit?usp=sharing
    % https://docs.google.com/drawings/d/1dGy8L1YzqmDvxjE5rtdUlhWeQMskkwwPESmdhkQCYBU/edit?usp=sharing
    \vspace{-7mm}
    \caption{\textbf{Augmentation with Obstacle Avoidance.} %\textcolor[HTML]{f4b371}{Top:} 
    Top: original demo;
    Middle: augmented trajectories without obstacle avoidance; 
    Bottom: augmented trajectories with obstacle avoidance.}
        
    \label{fig:method_obstacles}
    \vspace{-6mm}
\end{wrapfigure}

% - L_funnel
We preserve the original contact dynamics using a delta funnel loss $\mathcal{L}_\mathrm{funnel}$ to produce trajectories that converge consistently 
to the same pre-contact pose of the original demonstration. Let $R$ and $t$ represent the rotation and translation components of $x\in SE(3)$, we have % DB: do we need to save space? otherwise I'd make the loss terms and the weights definition equations (not inline)
$\mathcal{L}_{\mathrm{funnel}}
=\sum_{k} w_{k}\,\bigl\lVert t^{k}-t_{\mathrm{ee}}^{k}\bigr\rVert_{2}^{2}$, where $w_k$ is a temporally dependent weight defined as, $w_k = w_\mathrm{min} + (w_\mathrm{max}-w_\mathrm{min}) \cdot \big(\frac{k}{T}\big)^{3}, \sum_{k}w_k=1$, $0 \leq w_{\{\mathrm{max}, \mathrm{min}\}} \leq 1$. 
Novel view rendering from 3DGS suffers from floater artifacts and blurry scene reconstruction when the rendering viewpoint differs too much from the training viewpoint distribution. To free the generated data from these rendering artificats, which may have negative impact on downstream policy training, we introduce $\mathcal{L}_\mathrm{render}$ to optimize each generated pose $x^k$ to be close to the 6D pose distribution of the original demonstration and scanning views within a ball neighborhood $\mathcal{N}_k$,  $\mathcal{L}_\mathrm{render} = \sum_{k}\sum_{j\in\mathcal{N}_k} \Bigl\|\log\bigl((R^k)^\top R^{j}_{\mathrm{ee}}\bigr)\Bigr\|^2
+ \bigl\|t^k - t^{j}_{\mathrm{ee}}\bigr\|^2_2$. % DB: you specify a l2-norm for L funnel, but no norms here - what norm is used for the (hopefully close to) identity matrix in the first term? frobenius? -- also, this is the only term that ensures that the camera will be looking in the correct direction at contact; feels important but not mentioned
% - L_collision
To ensure the generated trajectories are free of collisions with the environment, we incorporate a collision-loss $\mathcal{L}_\mathrm{collision} = -\sum_{k}\mathrm{tsdf}(x^k)$. 
% - L_smooth --> DB: new lines for each loss term at least, if we don't do equations? otherwise this reads way too dense IMO.

In addition, to ensure smoothness of the generated trajectory, we introduce $\mathcal{L}_\mathrm{smooth}$ to penalize velocity jerkiness. For free-space augmentation, the trajectory initialization is produced by linear interpolation between the newly sampled $x_\mathrm{init}$ and the pre-contact demonstration pose $x_\mathrm{goal}$. 

\textbf{Obstacle Augmentation and Collision-Free Trajectory Generation.}
\label{sec:method-traj-opt-obstacle}
As shown in Fig.~\ref{fig:method_obstacles}, given an obstacle -- a point cloud sampled from an Objaverse~\cite{deitke2023objaverse} object, we compute convex hull of the object to update the scene TSDF. For obstacle-avoiding augmentation, an initial-guess trajectory is produced using RRT$^*$~\cite{karaman2011sampling}, sampling a trajectory that connects the new $x_\mathrm{init}$ and the pre-contact demo pose $x_\mathrm{goal}$, already avoiding the added obstacle in the scene. Then, we use the same trajectory optimization formulation above, with the added collision constraint, $X \cap \mathrm{convhull}(P_\mathrm{obstacle})= \emptyset$.

% To ensure that this trajectory is smooth and without unnecessary turns, we add a momentum term to the steering function in RRT to encourage sampling points to lie on the plane formed by $x_\mathrm{init}$, $x_\mathrm{goal}$ and the last selected point. 
% and physically-feasible demonstration poses will also allow our view generation module to produce visual data with good rendering quality with minimal floater artifacts,
% \begin{wrapfigure}{r}{0.4\textwidth}  
%     \centering
%     \vspace{-5mm}
%     \includegraphics[width=\linewidth]{figures/obstacle.pdf} % https://docs.google.com/drawings/d/1f0akTNsbzF0gdQgmobVUiCCFn2fVpb6GiuLbvFE5gZI/edit?usp=sharing
%     \vspace{-4mm}
%     \caption{\textbf{Augmentation with Obstacle Avoidance.} Top: The demonstrated and augmented trajectories (left), and their signed distance to the inserted obstacle (right). Bottom: The views from the demonstration without the obstacle and the augmented views from one new trajectory.}
%     \label{fig:method_obstacles}
% \end{wrapfigure}

% =====================================================
\subsection{View Generation via Fisheye 3D Gaussian Splatting}\label{sec:method-demo-view}
\label{sec:method-view-aug}
We generate the visual observations that correspond to the augmented (action) trajectories using \textit{Fisheye Gaussian Splatting}, % DB: again, is this a new name we want to coin? otherwise it might be "fisheye 3DGS" or smth.
both in free space and in scenes with added 3D obstacles. 
% (Sec~\ref{sec:method-view-aug-free-space})
% (Sec~\ref{sec:method-view-aug-obstacle})
We optimize the 6D trajectories from (Sec. \ref{sec:method-traj-opt}) to match the input video’s viewpoint distribution, yielding collision-free paths that maintain high-fidelity Gaussian Splat renderings by keeping views within the distributions of the training viewpoints for 3DGS.

\textbf{Free-Space View Generation.} We use the generated free-space trajectories that start from different initial scene observation directions to render from the trained Fisheye 3DGS of the scene to generate corresponding visual fisheye image observations.
\label{sec:method-view-aug-free-space} % DB: I don't think this paragraph is necessary -- all you say here is already mentioned in the paragraph before; is there anything more to say about this type of augmentation?

\textbf{Obstacle-Scene View Generation.} We augment the original Fisheye 3DGS of the scene with a trained Fisheye 3DGS % DB: you mean "Fisheye 3DGS reconstruction of the object"? a trained 3DGS may be unclear
of obstacles obtained from Objverse to generate unseen scene configurations. Then, we use the generated obstacle-avoiding trajectories starting from different initial scene observation directions to render from the trained Fisheye 3DGS of the scene to generate corresponding obstacle-avoiding visual fisheye image observations.
\label{sec:method-view-aug-obstacle} % DB: again, I feel like I already read all of this information before. I don't think this needs separate \textbf subheadings - just integrate any additional information into the first paragraph

\subsection{Action-View Augmentation for Visuomotor Policies}\label{sec:method-policy}
We train a Diffusion Policy~\cite{chi2023diffusion} on the union of the original and augmented datasets, \(\mathcal{D}\cup\tilde{\mathcal{D}}\), using a CLIP‐pretrained ViT‐B/16 encoder~\cite{radford2021learning,dosovitskiy2020image} with a relative action representation.

\textbf{Action-View Data Compilation.}
\label{sec:method-compile-demo-data}
We collect these original demonstrations $\mathcal{D}$ on a modified UMI~\cite{chi2024universal} with an iPhone. ARKit VIO provides metric 6D end-effector poses \(a_{ee}\), removing the need for an extra AprilTag SLAM mapping round. The gripper-opening width \(a_{gp}\) is measured via fiducial markers. For augmented data \(\tilde{\mathcal{D}}\), each fisheye observation \(o_{\mathrm{fish}}\) is segmented by SAM2~\cite{ravi2024sam} and we overlay the gripper onto rendered images, yielding \(\tilde o_{\mathrm{fish}}\). We assign trajectory-optimized 6D poses (Sec.~\ref{sec:method-traj-opt}) as \(\tilde a_{ee}\) and retain the original gripper-opening width as \(\tilde a_{gp}\).

\section{Experiments}
Our experiments in real-world and simulated environments aim to answer the following key questions: 1) Does action-view augmentation help imbue policies with improved robustness against unseen initial configurations and obstacles(\S \ref{sec:eval_1})? % DB: help what? there is a quantifier missing here - help success rate, help generalization?
2) How should augmentation be performed (\S \ref{sec:eval_2})? % DB: what does this refer to? free space vs obstacle? view only vs action vs combined? unclear
and 3) How much augmentation is beneficial (\S \ref{sec:eval_3}). % DB: how strong?
% DB: still not a fan of the paragraph symbol :P I'm sure there is a styleguide that asks for Sec. or Section being used instead.

% \begin{itemize}[leftmargin=3mm]
%     \item \textbf{Does action-view augmentation help?}  
%     We examine the impact of action-view augmentation in handling out-of-distribution configurations and unseen collision avoidance scenarios.   
    
%     \item \textbf{How should augmentation be performed?}
%     We compare alternative augmentation strategies, analyzing the effects of key design choices on performance.  

%     \item \textbf{How much augmentation is beneficial?}  
%     We study the trade-off between the extent of action-view augmentation and data quality, aiming to find an optimal balance.
% \end{itemize}

\begin{figure}[t]
    \centering
    % \includegraphics[width=0.46\linewidth]{example-image}
    % https://docs.google.com/drawings/d/1bKRQN2aG4OdJ0TTEuO-zIVeRxvVCDtSmSYEU94Ulk2Q/edit?usp=sharing
   \vspace{-5mm}
     \includegraphics[width=0.8\linewidth]{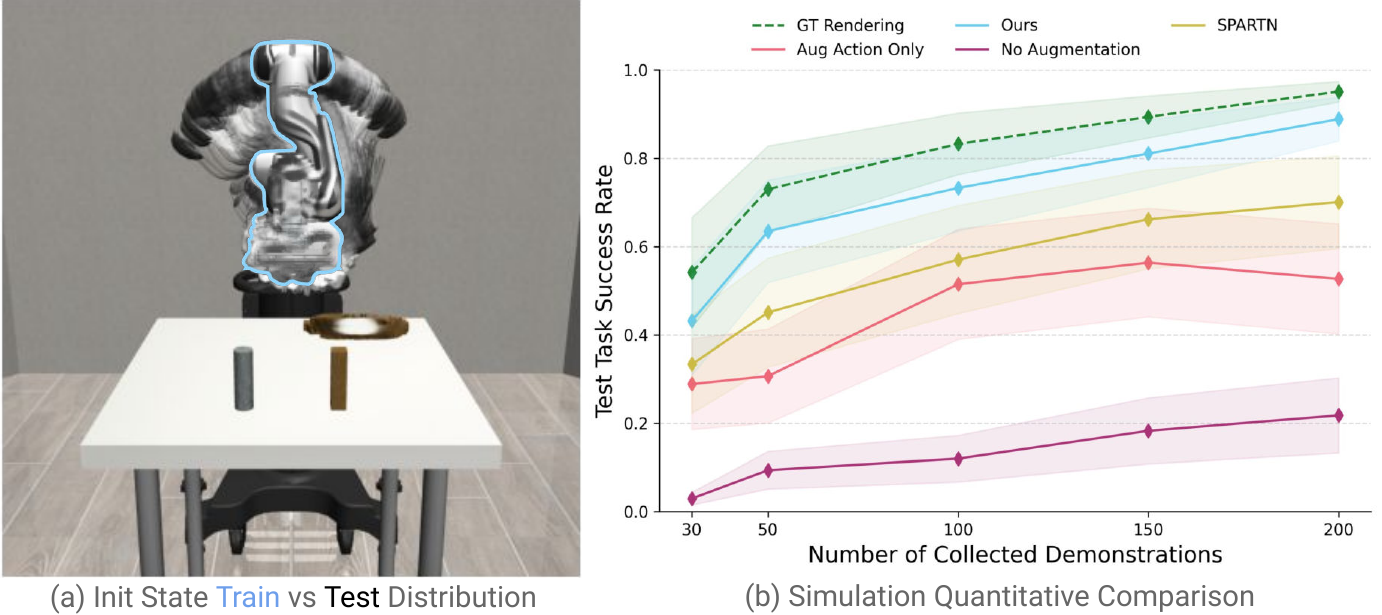}
    \caption{\textbf{Simulation Evaluation.} (a) Initial state distribution for training data highlighted in \textcolor{cyan}{blue} overlay over custom test data. (b) Task success rate with action-view augmentation, compared to no augmentation, oracle action-view augmentation and other augmentation baselines. % DB: do you define the oracle? if so, you call it differently in the text...
    % \todo{move the legend to the top, update the color for NERF too close to no augmentation, update the order to match ranking}
    }
   \vspace{-3mm}
    \label{fig:exp2-3}
\end{figure}

\noindent \textbf{Simulation Evaluation.}
For evaluation in simulation, we use the \textit{RoboMimic}~\cite{mandlekar2021matters} ``square'' task to evaluate the performance gain afforded by \ours~in free-space augmentation, compared to no augmentation, augmentation with ground truth novel-view rendering (obtainable in MuJoCo~\cite{todorov2012mujoco}), % DB: what is "perfect" novel view rendering? how do we get it? would trip me up as a reviewer
and representative baselines -- \texttt{Aug Action Only}~\cite{florence2019self}, and \texttt{SPARTN}~\cite{zhou2023nerf}(\S\ref{sec:eval_2}). We used the 200 expert-collected task demonstrations provided in RoboMimic as the base dataset. The RoboMimic ``square'' task is a peg-in-hole task that requires a Franka Emika Panda robot to pick a square nut and insert it onto a rod. To follow the same data format as in~\S\ref{sec:method-scene-reconstruction}, we convert the pinhole image observations from the wrist camera into Fisheye images using intrinsic \& distortion parameters of a GoPro Fisheye lens with a 155$^\circ$ FoV. To study the effects of different methods of augmentation, we keep the augmentation scheme constant -- for each base demonstration episode, we generate 20 free-space augmentation episodes. The initial camera pose is sampled within a $\theta$ quaternion cone around the initial camera pose of each base demonstration, with $0.15$m as the scaled radius of the quaternion sphere, and the initial camera viewing direction defined as the zero quaternion. For quantitative comparison, shown in Fig.~\ref{fig:exp2-3}(b), we found $\theta = 50 ^\circ$ to strike a balance between novel-view rendering quality and policy performance, as detailed in \S\ref{sec:eval_3}. % DB: is this the sweetspot for ours? for all? if only for one, is this fair for the other methods? opens the evaluation up for critique to be unspecific here
All the compared methods are used to train a Diffusion Policy~\cite{chi2023diffusion} on subsets of $\{30, 50, 100, 150, 200\}$ of the expert dataset, respectively, and tested on a fixed set of 1000 initial robot configurations. These are sampled with the end-effector poses within a quaternion cone of $50^\circ$ and $0.15$m radius with respect to the base RoboMimic dataset, while keeping the test object configurations unaltered. The distributions of training and testing initial states are shown in~Fig.\ref{fig:exp2-3}(a), with the resulting policy success rates reported in~Fig.\ref{fig:exp2-3}(b).

%  Because our test set covers a much wider range of initial camera poses than robomimic’s native train and test distribution, the base policy performs worse here than in the original benchmarks.
% \todo{describe and task, how many human demonstration it starts with? Make sure to describe the evaluation is OOD that's why base policy is so bad}

\begin{figure}[t]
    \centering
    \includegraphics[width=0.99\linewidth]{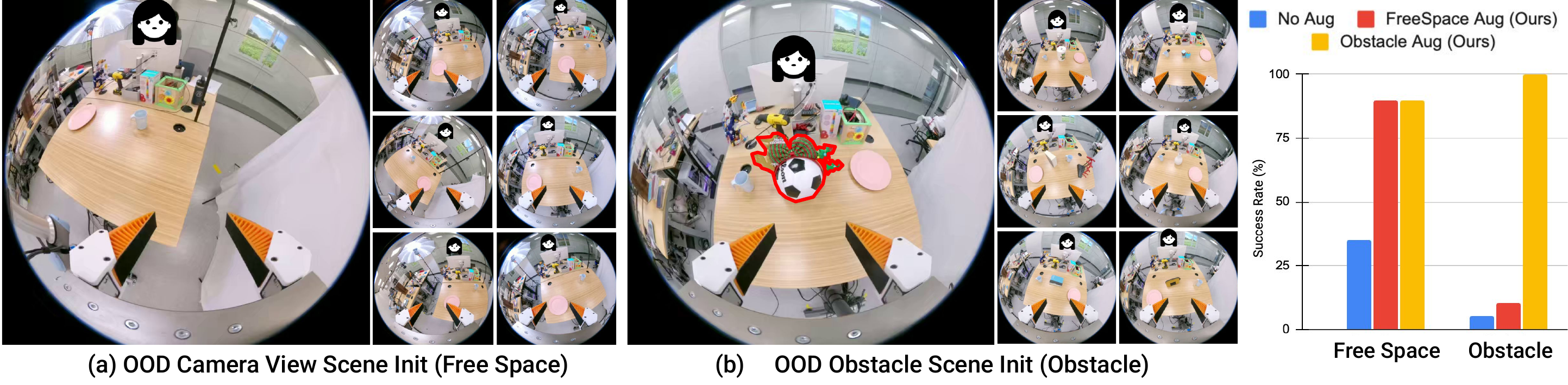}
    % https://docs.google.com/drawings/d/1bwY5k3UwlADAY42cudEukw94THjHGPERxyhGnAcgx44/edit?usp=sharing
    
    % \begin{tabular}{c|c|c}
    % \toprule
    %       Method & FreeSpace & Obstacle \\
    %       NoAug & 7/20  & 1/20 \\
    %       FreeSpace  & 18/20 & 2/20 \\ 
    %       ObstacleAug & 18/20 &  20/20 \\
    % \bottomrule 
    % \end{tabular}
    \vspace{-2mm}
    \caption{\textbf{Real-world Evaluation.} We report task performance for two versions of our augmented policies -- trained with free-space augmentation (\textit{FreeSpace Aug}), and free-space \& obstacle-distractor augmentation (\textit{Obstacle Aug}) -- against a vanilla policy trained with no augmentation (\textit{No Aug}). Initial states for a subset of all evaluation episodes for (a) OOD camera view test case, (b) OOD obstacle distractor test case shown. (c) Success rates, averaged over 20 evaluation episodes.}
    \vspace{-5mm}
    % \todo{Left show train test distribution (free space) Middle (obstacle) Right show number, maybe as a bar chart}
    % \caption{\textbf{Results in Realworld Environment.} \todo{Left show train test distribution (free space) Middle (obstacle) Right show number, maybe as a bar chart}}
   \label{tab:realworld}
\end{figure}

\noindent \textbf{Real-world Evaluation.} 
In our real-world evaluation, we aim to determine the effectiveness of \ours~for enabling visuomotor policies to handle out-of-distribution (OOD) scenarios with respect to our UMI-collected training data distribution on the following two axes: (a) OOD robot and object initial states; (b) unseen obstacles in the scene. We report evaluation results on a cup serving task. This task requires the robot to pick a cup with its handle to the left and place it on the pink serving plate, as shown in Fig.~\ref{tab:realworld}. We define the task as successfully completed when the cup is placed upright on the serving plate with its handle within $\pm10^{\circ}$ towards the left of the table. The base dataset includes 89 demonstration episodes, collected by a single demonstrator using UMI~\cite{chi2024universal}. All demonstrations were collected in obstacle-free scenes, with all initial camera views in an upright overhead orientation (Fig.~\ref{fig:teaser}).

We manually select 50 episodes that produce the best novel‐view rendering results for both free‐space and obstacle augmentation. % DB: how? manually? automatically?
From these, we generate 7245 free‐space augmented episodes by sampling initial camera orientations within a $45^\circ$ quaternion cone around each original viewpoint, with the Euclidean distance between the initial demonstration start pose and the pre-contact pose as the radius of the quaternion sphere. For obstacle augmentation, we select 50 objects from Objaverse~\cite{deitke2023objaverse}, render 256 orbit views for each to train a fisheye-3DGS per obstacle. Then, by integrating each obstacle into the original demonstrations' 3DGS scenes, we generate 5000 obstacle-aware episodes in total. We train Diffusion Policy~\cite{chi2023diffusion}, following UMI's policy interface protocols, to obtain a \texttt{No Aug} policy with the original human-collected data, a \texttt{FreeSpace Aug} policy with the original data and additionally generated free-space augmentation data, and an \texttt{Obstacle Aug} policy with the original data and additionally generated obstacle augmentation data. Policy success rates are averaged over 20 evaluation episodes for two test scenarios: (a) OOD camera‐view and (b) OOD obstacle initialization, are shown in~Fig.~\ref{tab:realworld}, with a subset of the init distributions shown. We further evaluated the three policies in a harder obstacle setup with more challenging placements and larger, more complex shapes, as shown in~\ref{sec:eval_challenge}.

%  The following sections discuss each key question one by one. 

% \todo{describe real-world environment and task}
% Fig. \ref{fig:exp2-3} summarizes the quantitative results in simulation. Fig. \ref{tab:realworld} summarizes the quantitative results in real world. 

% \begin{figure}[t]
%     \centering
%     % \includegraphics[width=0.2\linewidth]{example-image}›
%     % \includegraphics[width=0.2\linewidth]{example-image}
%     \includegraphics[width=0.8\linewidth]{figures/real_world_exp_1.pdf}
%         \begin{tabular}{c|c|c}
%     \toprule
%           Method & FreeSapce & Obstacle \\
%           NoAug & 7/20  & 1/20 \\
%           FreeSpace  & 18/20 & 2/20 \\ 
%           ObstaleAug & 18/20 &  20/20 \\
%     \bottomrule 
%     \end{tabular}
%     \caption{\textbf{Results in Realworld Environment.} \todo{Left show train test distribution (free space) Middle (obstacle) Right show number, maybe as a bar chart}}
%    \label{tab:realworld}
% \end{figure}

\subsection{Does Action-View Augmentation help?} 
\label{sec:eval_1}
\noindent \textbf{Action-view augmentation helps in improving sample efficiency.} 
% In Fig~\ref{fig:exp2-3}(b), we explore the lower and upper bounds on policy improvement achievable by \ours~in simulation. We evaluate the performance improvement achieved by our action-view augmentation scheme compared to: (1) purely human-collected demonstrations without augmentation (Fig~\ref{fig:exp2-3}, No Augmentation), and (2) a combination of human-collected demonstrations and augmented demonstrations generated with ground-truth rendering (Fig~\ref{fig:exp2-3}, GT Rendering), serving as an oracle to establish an upper bound for our method in the limit of perfect novel-view synthesis.
To investigate how much improvement our augmentation scheme provides compared to no augmentation, we evaluate visuomotor policies trained solely on the converted fisheye RoboMimic expert dataset, totaling 200 episodes, without any augmentation. Conversely, to establish an upper bound for our action-view augmentation scheme if it would provide perfect novel-view synthesis, we train a policy using the union of fisheye RoboMimic expert demonstrations and free-space augmentation demonstrations generated using \ours~with ground-truth rendering (which is only possible in simulation). This produces an oracle policy to provide an upper bound for our action-view augmentation scheme. % DB: read this and the paragraph above; there is some (word-by-word) overlap here - try not to repeat information
Figure~\ref{fig:exp2-3}(b) shows that, with the same amount of original expert demonstrations, \textcolor[HTML]{6dcae6}{our} action‐view augmentation closely tracks the perfect \textcolor[HTML]{1f8a3c}{GT‐rendering} upper bound, with a performance gap of 8\% in the low‐data regime and 11\% in the high‐data regime. Compared to policies trained \textcolor[HTML]{a83579}{without augmentation}, % DB: does this refer to No Augmentation? then I'd say purple, no? I was confused if you mean the second-to-last (Aug Action only)
our free space augmentation provides an average of 56\% task success rate improvement. % DB: average over these 5 datapoints, right? maybe add "success rate improvement; from ??% to ??%." to make this point clear - it's a big improvement afterall, so highlight it!
In the real-world experiment shown in Fig~\ref{tab:realworld}, we find that, with the same amount of human demonstrations, our free space augmentation provides a performance boost of 55\% over policies trained without augmentation. % DB: again, exact numbers would be good imo. here, you could/should actually put the numbers into the barchart (there's an option in google sheets to add the data above the corresponding bar)

% \todo{real world results, no aug, vs free space aug}
% \todo{Describe GT rendering result, and compare ours with no augmentation.
% Then describe our method result both in simulation and realworld.  With the same amount of training data how much more performence improvement we can get}  

\noindent \textbf{Action-view augmentation helps in obstacle avoidance.}  
In the real-world experiment, shown in Fig.~\ref{tab:realworld}, we find that, by augmenting obstacle-free demos with our generated obstacle-avoiding demos, % DB: these sentences (also above) are a bit too nested imo
our action-view augmentation equips the visuomotor policy to robustly complete the task while avoiding obstacles -- behaviors not shown in the original human demonstrations. We find that our full \texttt{Obstalce Aug} is able to complete the task with a 100\% success rate, significantly outperforming \texttt{FreeSpace Aug} with 10\% and \texttt{No Aug} with 5\% success rates, respectively.
% \todo{1. point to the qualitative results of obstacle avoidance
% 2. summarize the quantitative result in realworld}

% \begin{figure*}[t]
%     \centering
%     \includegraphics[width=\linewidth]{figures/ablation.pdf} % https://docs.google.com/drawings/d/1l4iqUmPgafQH70mSZnre0R__rCRHSciFroa3V-9RA2o/edit?usp=sharing
%     \caption{\textbf{How to augment?} Qualitative comparison between different augumentation methods. Left: IDs. Mid and right: SDF loss. -- TODO: more columns for more loss terms? -- rows with the generated views (per-id-color as border)}
%     \label{fig:tradeoff}
% \end{figure*}

\subsection{How to Augment?}
\label{sec:eval_2}
\noindent\textbf{Comparison to action-only augmentation.} One simple and effective augmentation for local feedback stabilization is \texttt{Aug Action Only}~\cite{florence2019self}, which slightly perturbs proprioceptive and gripper action data while leaving visual data unchanged. While effective for small, local robot state variations under third‐person views, it breaks down with eye-in-hand observations, where minor end-effector pose changes produce drastic visual shifts. We replicate this baseline by applying free-space augmentation in the end-effector and gripper actions, but using the original visual observations. As Fig.~\ref{fig:exp2-3} shows, \texttt{Aug Action Only} boosts the average success rate by 29\% over \texttt{No Aug} -- peaking at 56\%. In turn, \texttt{Ours} outperforms \texttt{Aug Action Only} by up to 35\%, especially in high-data regimes.

% \todo{briefly describe why this method makes sense, and describe its result -- indeed better than no augmentation, and describe its issue. 
% Finally compared with our method how much worse xx point to table and figure. }

\noindent\textbf{Comparison to single-step augmentation.}
In this experiment, we compare our proposed method, which augments both visual and action data for the whole trajectory, with prior work SPARTN~\cite{zhou2023nerf}, which augments visual and action data for a single step and uses NeRF to reconstruct the visual scene. % DB: again, very nested. break this up into at least 2 sentences with an easier structure, e.g., "we compare ours with spartn. while ours augments ..., this baseline uses ..."
Single-step action-view augmentation using \texttt{SPARTN} improves policies' performance for OOD camera views, as shown in Fig.~\ref{fig:exp2-3} with a performance gain of 41\% over \texttt{No Aug}. 
% However, the single-step augmentation nature of SPARTN prevents the augmented data from containing smooth, collision-avoiding behaviors that can only be controlled on the trajectory level, leading to policies that more easily end up in unseen configurations for which no recovery behavior is present in the training data. 
However, SPARTN’s single‐step augmentation cannot produce smooth, trajectory‐level collision‐avoidance behaviors, more easily causing policies to enter unseen configurations for which no recovery behaviors exist in the training data. % DB: under what condition - this is unclear; rephrased these sentences to cut down on complexity
By comparison, \ours~performs trajectory-level action-view augmentation. We observe that \texttt{Ours} thereby is able to outperform \texttt{SPARTN}, boosting the average success rate by 15\% -- peaking at an improvement of 18\%. 
more easily end up in unseen configurations for which no recovery behavior is present in the training data.
\subsection{How Much to Augment?} 
\label{sec:eval_3}
While larger rotation bounds increase diversity, they can harm rendering quality under limited viewpoint coverage. To quantify this trade-off, we trained visuomotor policies on 50 RoboMimic ``square'' export demonstrations (fisheye, eye-in-hand), augmenting each with rotation bounds of \(\{20^\circ,30^\circ,40^\circ,50^\circ,60^\circ\}\), generating 20 augmented episodessamples per demo. We then evaluated on a held-outout test set shown in Fig~\ref{fig:exp2-3}. As shown in Fig~\ref{fig:tradeoff}, success rates plateaued at \(50^\circ\), which we therefore adopt for all other experiments.

\section{Conclusion}
% In this work, we present~\ours~, an offline data augmentation framework for visuomotor polices that enables manipulation polices to acquire skills not demonstrated from the original dataset: avoiding obstacles and handle diverse out-of-distribution initial configurations of unseen robots in the real world. By utilizing existing manipulation demonstrations with limited initial configuration coverage and obstacle-free scenes, through scene reconstruction and novel view rendering, we generate visually realistic observation data from diverse viewpoints, and we use trajectory optimization to generate physically feasible, collision-free action trajectory data. By training policies with these generated augmentation data, we imbue robot polices with improved robustness against diverse out-of-distribution initial camera views and the ability to gracefully avoid obstacles while completing the original task. 

We present~\ours, an offline data-augmentation framework for visuomotor policies that effectively endows robots with skills not demonstrated in original demo -- obstacle avoidance and robustness to novel initial robot configurations, by generating visually realistic and physically feasible \textit{trajectory-level}, \textit{obstacle-avoiding}, \textit{eye-in-hand}, \textit{fisheye} action-view demonstrations.
% Based on obstacle-free demonstrations with limited viewpoint coverage, we first reconstruct each scene and render realistic observations from diverse camera poses. We then apply constrained trajectory optimization to generate physically feasible, collision-free action sequences. Training on this augmented dataset endows policies with the ability to complete the original task from out-of-distribution views and to gracefully avoid obstacles at deployment.

% By offloading the data collection effort from human time to compute time, we significantly reduce the human effort needed in clock-time data collection and data iteration, and provide precise control over the data generated, vastly expediting the data collection iteration cycle for visuomotor policy deployment. 

\clearpage

% The acknowledgments are automatically included only in the final and preprint versions of the paper.
\acknowledgments{This work was supported in part by the Toyota Research Institute, NSF Award \#2143601, \#2037101, and \#2132519. The views and conclusions contained herein are those of the authors and should not be interpreted as necessarily representing the official policies, either expressed or implied, of the sponsors.}

%===============================================================================

\section{Limitations \& Future Work}
With only a single eye-in-hand moving camera, the view coverage of the demonstration stage is inadequate for dynamic scene reconstruction or for generating novel views far from the original viewpoints. As a result, we currently restrict the \ours~pipeline to static scenes before or after contact. Future work could explore more advanced sensing setup, such as multi-camera rigs or ToF sensors, or adopt advanced dynamic reconstruction methods that demand fewer training viewpoints~\cite{zhu2024fsgs, chung2024depth}. 

Our novel-view generation module inherits 3DGS’s multi-view inconsistent nature, yielding floating artifacts for generated viewpoints outside of the training viewpoint distribution. This could be alleviated by adopting inherently view-consistent representations like 2DGS~\cite{huang20242d}.

Similar to UMI~\cite{chi2024universal}, since the kinematic limits of the downstream deployment robots are unknown at the time of data collection, the generated demonstration trajectories do not account for kinematic limits of the downstream deployment robots. We carefully bound both, the sampling range for initial poses and the placement of obstacles, to ensure that the generated trajectories lie within the task space of the deployment robot. Our work could be extended to incorporate the downstream robot kinematics constraint in the trajectory optimization module, ensuring embodiment-aware trajectory generation with respect to the specific deployment robot and enable kinematically feasible and smooth action trajectory, not just in the task space as addressed by this work, but also in the configuration space. Furthermore, this could allow retrofitting the original embodiment-agnostic demonstrations to be kinematically feasible for the downstream robot hardware deployment, thereby allowing an embodiment-aware policy learning framework that can transfer skills from semantically and physically valid but hardware-infeasible actions to different robot embodiments.

\bibliography{reference}  % .bib
\newpage
\appendix
\renewcommand{\thesection}{A.\arabic{section}}
\renewcommand{\thefigure}{A\arabic{figure}}
\renewcommand{\thetable}{A\arabic{table}}
\setcounter{section}{0}
\setcounter{figure}{0}
\setcounter{table}{0}

\section{How Much to Augment?}
\begin{figure}[h]
    \centering
    \includegraphics[width=0.5\linewidth]{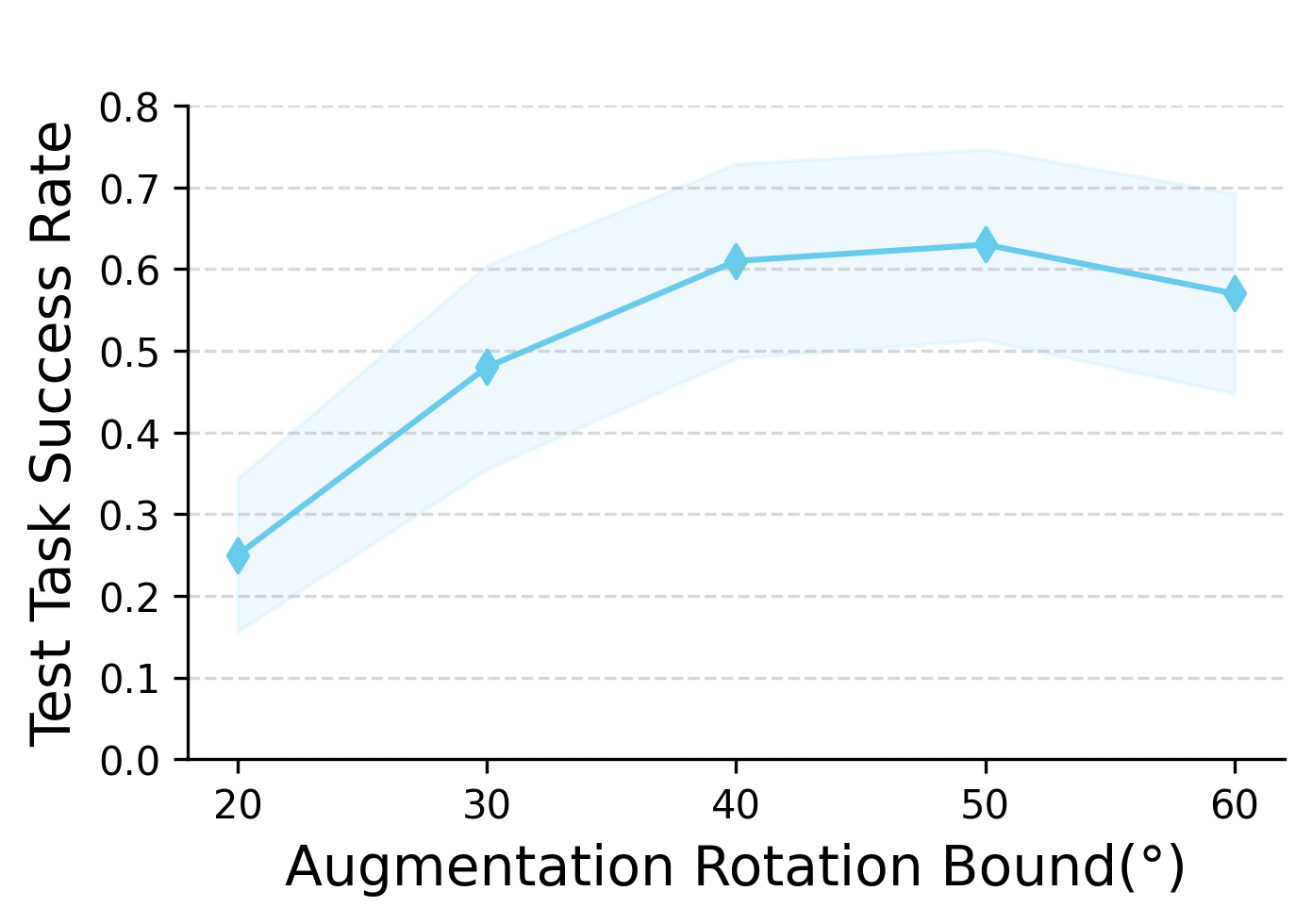}
    \caption{\textbf{How much to augment?} While larger augmentation range could increase the data diversity, it also reduces the image rendering quality due to limited demonstration viewpoint coverage. We found \(50^\circ\) % DB: 50deg of what? wrt to what? 50deg celsius :)
    as the optimal trade-off.}
    \label{fig:tradeoff}
\end{figure}

\section{Real World Evaluation on Challenging Obstacles.}
\label{sec:eval_challenge}
\begin{figure}[h]
    \centering
    \includegraphics[width=1\linewidth]{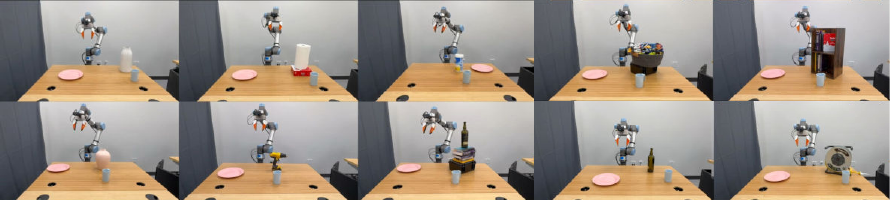}
    % https://docs.google.com/drawings/d/1bwY5k3UwlADAY42cudEukw94THjHGPERxyhGnAcgx44/edit?usp=sharing
    
    % \begin{tabular}{c|c|c}
    % \toprule
    %       Method & FreeSpace & Obstacle \\
    %       NoAug & 7/20  & 1/20 \\
    %       FreeSpace  & 18/20 & 2/20 \\ 
    %       ObstacleAug & 18/20 &  20/20 \\
    % \bottomrule 
    % \end{tabular}
    \vspace{-2mm}
    \caption{\textbf{Challenging Obstacle Evaluation.} Initial states for all 10 evaluation episodes for \textit{Challenging Obstacle} experiment. Please see the accompanying video for more comparisons.}
    \vspace{-5mm}
   \label{fig:realworld_challenging}
\end{figure}

To further examine the capability of our policies enabling obstacle avoidance enhancement, we additionally evaluated policy performance for the cup serving task on a set of more challenging scenarios with more challenging obstacle placement, and obstacles of larger size and more diverse geometric shape; we term this set of experiments \textit{Challenging Obstacles}. As shown in Tab~\ref{tab:challenging_obstacle}, we conducted 10 trials on 10 different obstacle sets as shown in Fig.~\ref{fig:realworld_challenging}, on the same three policies \texttt{No Aug}, \texttt{FreeSpace Aug}, \texttt{Obstacle Aug}, as tested in real world experiments for \textit{Free Space} and \textit{Obstacle} as reported in manuscript, and found that ours \texttt{Obstalce Aug} was able to complete 10/10 trials, while \texttt{No Aug} and \texttt{FreeSpace Aug} both fail complete any trials.

\begin{table}[h]
  \centering
  \begin{tabular}{l|c}
    \toprule
    Method                   & Task Success Rate \\
    \midrule
    \texttt{No Aug}                   & 0\%                 \\
    \texttt{FreeSpace Aug} (Ours)     & 0\%                 \\
    \texttt{Obstacle Aug} (Ours)      & \textbf{100\%}        \\
    \bottomrule
  \end{tabular}
  \caption{\textbf{Real World Evaluation Results for \textit{Challenging Obstalces}}. Task success rate reported over 10 trials. }
  \label{tab:challenging_obstacle}
\end{table}

\end{document}